# X-TREPAN: A MULTI CLASS REGRESSION AND ADAPTED EXTRACTION OF COMPREHENSIBLE DECISION TREE IN ARTIFICIAL NEURAL NETWORKS

Awudu Karim[1], Shangbo Zhou[2]

College of Computer Science, Chongqing University, Chongqing, 400030, China.
awudubody@yahoo.com

## ABSTRACT

*In this work, the TREPAN algorithm is enhanced and extended for extracting decision trees from neural networks. We empirically evaluated the performance of the algorithm on a set of databases from real world events. This benchmark enhancement was achieved by adapting Single-test TREPAN and C4.5 decision tree induction algorithms to analyze the datasets. The models are then compared with X-TREPAN for comprehensibility and classification accuracy. Furthermore, we validate the experimentations by applying statistical methods. Finally, the modified algorithm is extended to work with multi-class regression problems and the ability to comprehend generalized feed forward networks is achieved.*

## 1. INTRODUCTION

Artificial neural networks are modeled based on the human brain architecture. They offer a means of efficiently modeling large and complex problems in which there are hundreds of independent variables that have many interactions. Neural networks generate their own implicit rules by learning from examples. Artificial neural networks have been applied to a variety of problem domains [1] such as medical diagnostics [2], games [3], robotics [4], speech generation [5] and speech recognition [6]. The generalization ability of neural networks has proved to be superior to other learning systems over a wide range of applications [7].

However despite their relative success, the further adoption of neural networks in some areas has been impeded due to their inability to explain, in a comprehensible form, how a decision has been arrived at. This lack of transparency in the neural network's reasoning has been termed the Black Box problem. Andrews et al. [8] observed that ANNs must obtain the capability to explain their decision in a human-comprehensible form before they can gain widespread acceptance and to enhance their overall utility as learning and generalization tools. This work intends to enhance TREPAN to be able to handle not only multi-class classification type but also multi-class regression type problems. And also to demonstrate that X-TREPAN can understand and analyze generalized feed forward networks (GFF). TREPAN is tested on different datasets and best settings for TREPAN algorithm are explored based on database type to generate heuristics for various problem domains. The best TREPAN model is then compared to the baseline C4.5 decision tree algorithm to test for accuracy.

Neural networks store their "Knowledge" in a series of real-valued weight matrices representing a combination of nonlinear transforms from an input space to an output space. Rule extraction attempts to translate this numerically stored knowledge into a symbolic form that can be readily comprehended. The ability to extract symbolic knowledge has many potential advantages: the knowledge obtained from the neural network can lead to new insights into patterns and dependencies within the data; from symbolic knowledge, it is easier to see which features of the data are the most important; and the explanation of a decision is essential for many applications, such as safety critical systems. Andrews et al. and Ticke et al. [9], [10] summarize several proposed approaches to rule extraction. Many of the earlier approaches required a specialized neural network architectures or training schemes. This limited their applicability; in particular they cannot be applied to in situ neural networks. The other approach is to view the extraction process as learning task. This approach does not examine the weight matrices directly but tries to approximate the neural network by learning its input-output mappings. Decision trees are a graphical representation of a decision process. The combination of symbolic information and graphical presentation make decision trees one of the most comprehensible representations of pattern recognition knowledge.

## 2. BACKGROUND AND LITERATURE REVIEW

### 2.1 Artificial Neural Network

Artificial neural networks as the name implies are modeled on the architecture of the human brain. They offer a means of efficiently modeling large and complex problems in which there may be hundreds of independent variables that have many interactions. Neural networks learn from examples by generating their own implicit rules. The generalization ability of neural networks has proved to be equal or superior to other learning systems over a wide range of applications.

### 2.2 Neural Network Architecture

A neural network consists of a large number of units called processing elements or nodes or neurons that are connected on a parallel scale. The network starts with an input layer, where each node corresponds to an independent variable. Input nodes are connected to a number of nodes in a hidden layer. There may be more than one hidden layer and an output layer. Each node in the hidden layer takes in a set of inputs (X1, X2, …, Xm), multiplies them by a connection weight (W1, W2, …, Wm), then applies a function, f(WTX) to them and then passes the output to the nodes of the next layer. The connection weights are the unknown parameters that are estimated by an iterative training method to indicate the connection's strength and excitation. The calculation of the final outputs of the network proceeds layer by layer [11]. Each processing element of the hidden layer computes its output as a function of linear combination of inputs from the previous layer plus a bias. This output is propagated as input to the next layer and so on until the final layer is reached. Figure 1 shows the model of a single neuron [12]

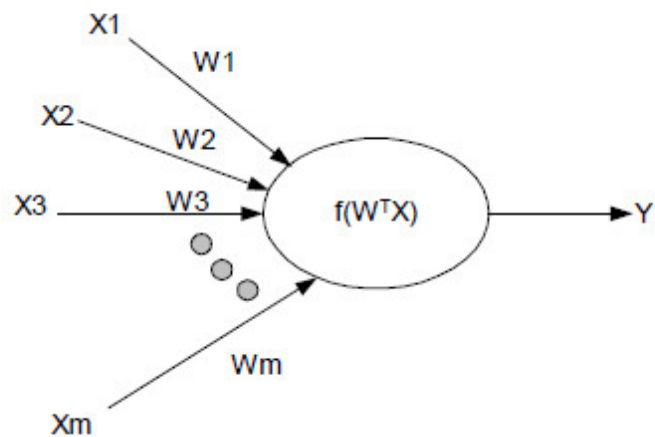

Figure 1. Model of a Single Neuron

The output of the neuron can be expressed as

$$Y = f\left(\sum_{i=1}^{m} w_i x_i\right), \text{ or}$$

$$Y = f(W^T X)$$

In the above equations, W is the weight vector of the neural node, defined as

$$W = [w_1, w_2, w_3, \ldots\ldots\ldots\ldots w_m]^T$$ and X is the input vector, defined as

$$X = [x_1, x_2, x_3, \ldots\ldots\ldots\ldots x_m]^T$$

Figure 2. Shows a typical neural network architecture representation.

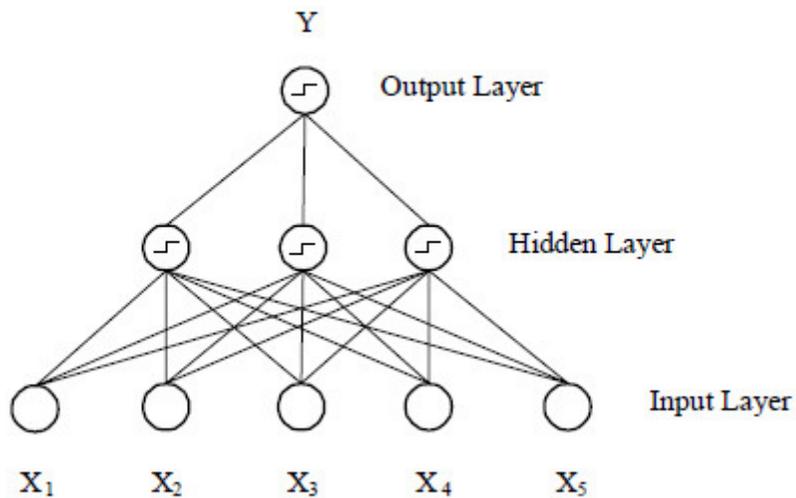

Figure 2. Neural Network Architecture

These are different types of activation functions that can be applied at the node of the network. Two of the most commonly used neural network functions are the hyperbolic and logistic (or sigmoid) functions. They are sometimes referred to as "squashing" functions since they map the inputs into a bounded range. Table 1 shows a list of activation functions that are available for use in neural networks

Table 1. Activation Functions used in Neural Networks Adapted from [13]

| Functions | Definition | Range |
|---|---|---|
| Identity | $x$ | $(-\infty, +\infty)$ |
| Logistic | $\dfrac{1}{(1-e^{-x})}$ | $(0, +1)$ |

| Hyperbolic | $\dfrac{e^x - e^{-x}}{e^x + e^{-x}}$ | $(-1,+1)$ |
|---|---|---|
| Exponential | $e^x$ | $(0,+\infty)$ |
| Softmax | $\dfrac{e^{-x}}{\sum_i e^{x_i}}$ | $(0,+1)$ |
| Unit Sum | $\dfrac{x}{\sum_i x_i}$ | $(0,+1)$ |
| Square root | $\sqrt{x}$ | $(0,+\infty)$ |
| Sine | $Sin(x)$ | $(0,+1)$ |
| Ramp | $\begin{cases} -1, x \leq -1 \\ x, -1 < x \\ +1, x \geq +1 \end{cases}$ | $(-1,+1)$ |
| Step | $\begin{cases} 0, x < 0 \\ +1, x \geq 0 \end{cases}$ | $(0,+1)$ |

## 2.3 Multilayer Perceptrons

Multilayer Perceptrons (MLOs) are layered feed forward networks typically trained with back propagation. These networks have been used in numerous applications. Their main advantage is that they are easy to use, and that they can approximate any input/output map. A major disadvantage is that they train slowly, require lots of training data (typically three times more training samples then network weights)[14].

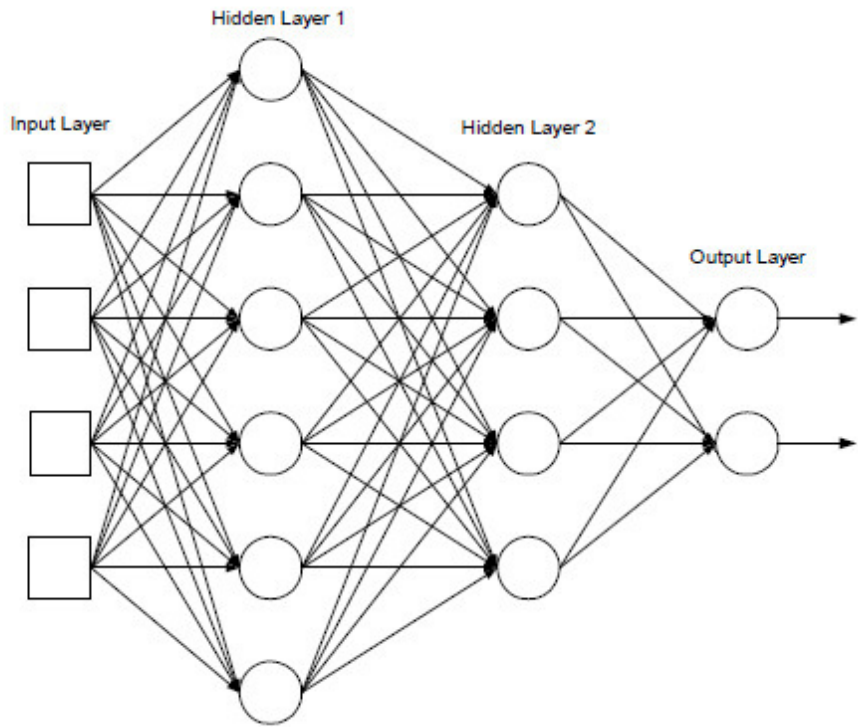

Figure 3. A schematic Multilayered Perceptron Network

A Generalized Feed Forward (GFF) network is a special case of a Multilayer Perception wherein connections can jump over one or more layers. Although an MLP can solve any problem that a GFF can solve, in practice, a GFF network can solve the problem more efficiently [14]. Figure 4 shows a general schematic of a Generalized Feed Forward Network.

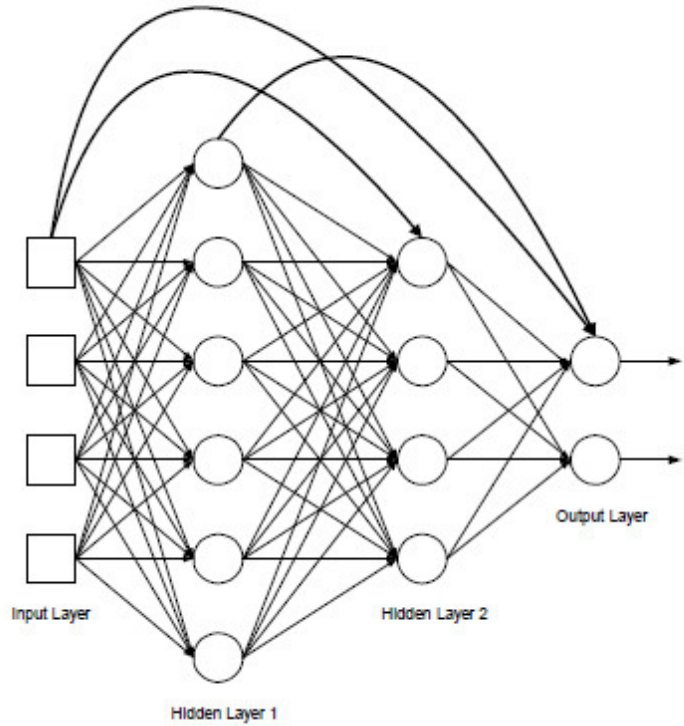

Figure 4. Generalized Feed Forward Networks

## 2.4 Neural Networks for Classification and Regression

Neural networks are one of the most widely used algorithms for classification problems. The output layer is indicative of the decision of the classifier. The cross entropy error function is most commonly used in classification problems in combination with logistic or soft max activation functions. Cross entropy assumes that the probability of the predicated values in a classification problem lie between 0 and 1. In a classification problem each output node of a neural network represents a different hypothesis and the node activations represent the probability that each hypothesis may be true. Each output node represents a probability distribution and the cross entropy measures calculate the difference between the network distribution and the actual distribution [15]. Assigning credit risk (good or bad) is an example of a neural network classification problem. Regression involves prediction the values of a continuous variable based on previously collected data. Mean square error is the function used for computing the error in regression networks. Projecting the profit of a company based on previous year's data is regression type neural network problem.

## 2.5 Neural Network Training

The neural network approach is a two stage process. In the first stage a generalized network that maps the inputs data to the desired output using a training algorithm is derived. The next stage is the "production" phase where the network is tested for its generalization ability against a new set of data.

Often the neural network tends to over train and memorizes the data. To avoid this possibility, a cross-validation data set is use. The cross validation data set is a part of the data set which is set aside before training and is used to determine the level of generalization produced by the training set. As training processes the training error drops progressively. At first the cross validation error decreases but then begins to rise as the network over trains. Best generalization ability of the network can be tapped by stopping the algorithm where the error on the cross validation set starts to rise. Figure 5 illustrates the use of cross-validation during training.

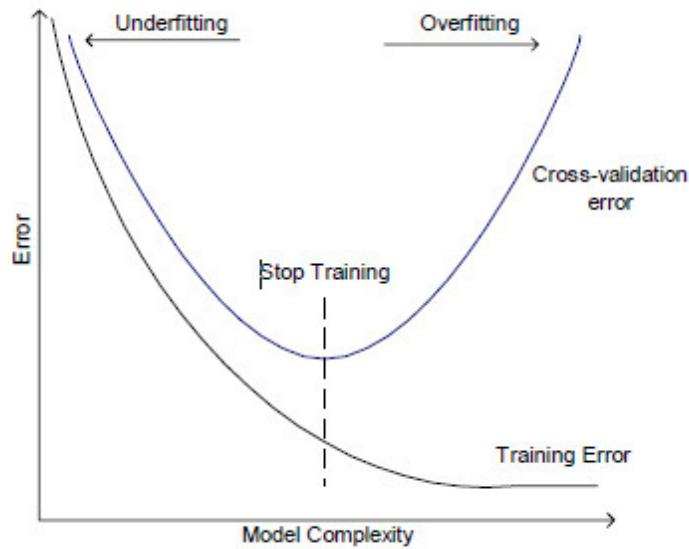

Figure 5. Use of cross-Validation during Training

## 2.6 Rule Extraction from Neural Networks

Although neural networks are known to be robust classifiers, they have found limited use in decision-critical applications such as medical systems. Trained neural networks act like black boxes and are often difficult to interpret [16]. The availability of a system that would provide an explanation of the input/output mappings of a neural network in the form of rules would thus be very useful. Rule extraction is one such system that tries to elucidate to the user, how the neural network arrived at its decision in the form of if-then rules.

Two explicit approaches have been defined to date for transforming the knowledge and weights contained in a neural network into a set of symbolic rules de-compositional and pedagogical [17]. In the de-compositional approach the focus is on the extracting rules at an individual hidden and/or output level into a binary outcome. It involves the analysis of the weight vectors and biases associated with the processing elements in general. The subset [18] algorithm is an example of this category. The pedagogical approach treats neural networks like black boxes and aims to extract rules that map inputs directly to its output. The Validity Interval Analysis (VIA) [19] proposed by Thrum and TREPAN [20] is an example of one such technique .Andrews et al [21] proposed a third category called eclectic which combines the elements of the basic categories.

## 2.7 Decision Trees

A decision tree is a special type of graph drawn in the form of a tree structure. It consists of internal nodes each associated with a logical test and its possible consequences. Decision trees are probably the most widely used symbolic learning algorithms as are neural networks in the non-symbolic category.

## 2.8 Decision Tree Classification

Decision trees classify data through recursive partitioning of the data set into mutually exclusive subsets which best explain the variation in the dependent variable under observation[22][23]. Decision trees classify instances (data points) by sorting them down the tree from the root node to some leaf node. This lead node gives the classification of the instance. Each branch of the decision tree represents a possible scenario of decision and its outcome.

Decision tree algorithms depict concept descriptions in the form of a tree structure. They begin learning with a set of instances and create a tree structure that is used to classify new instances. An instance in a dataset is described by a set of feature values called attributes, which can have either continuous or nominal values. Decision tree induction is best suitable for data where each example in the dataset is described by a fixed number of attributes for all examples of that dataset. Decision tree methods use a divide and conquer approach. They can be used to classify an example by starting at the root of the tree and moving through it until a leaf node is reached, which provides the classification of the instance.

Each node of a decision tree specifies a test of some attribute and each branch that descends from the node corresponds to a possible value for this attribute. The following example illustrates a simple decision tree.

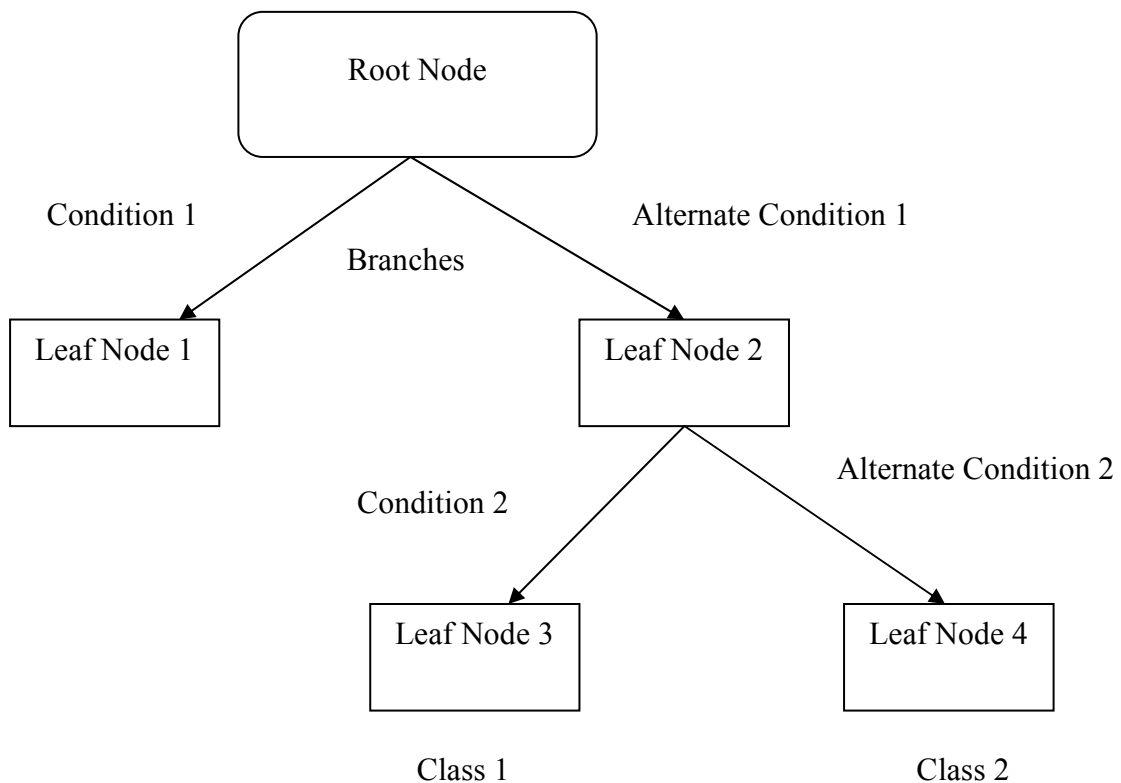

## 3. TREPAN Algorithm

The TREPAN [24] and [25] algorithms developed by Craven et al are novel rule-extraction algorithms that mimic the behavior of a neural network. Given a trained Neural Network, TREPAN extracts decision trees that provide a close approximation to the function represented by the network. In this work, we are concerned with its application to trained variety of learned models as well. TREPAN uses a concept of recursive partitioning similar to other decision tree induction algorithms. In contrast to the depth-first growth used by other decision tree algorithms, TREPAN expands using the best first principle. Thus node which increases the fidelity of the fidelity of the tree when expanded is deemed the best.

In conventional decision tree induction algorithms the amount of training data decreases as one traverses down the tree by selecting splitting tests. Thus there is not enough data at the bottom of the tree to determine class labels and is hence poorly chosen. In contrast TREPAN uses an 'Oracle' to answer queries, in addition to the training samples during the inductive learning process. Since the target here is the function represented by the neural network, the network itself is used as the 'Oracle'. This learning from larger samples can prevent the lack of examples for the splitting tests at lower levels of the tree, which is usually a problem with conventional decision tree learning algorithms. It ensures that there is a minimum sample of instances available at a node before choosing a splitting test for that node where minimum sample is one of the user specified parameters. If the number of instances at the node, say m is less than minimum sample then TREPAN will make membership queries equal to (minimum sample m) from the 'Oracle' and then make a decision at the node. The following illustrates a pseudocode of the TREPAN algorithm [26].

**Algorithm : TREPAN**

Input: Trained neural network; training examples $\{X_i, Y_i\}$ = where $y_i$, is the class label predicted by the trained neural network on the training example $X_i$, global stopping criteria.

Output : extracted decision tree

Begin

    Initialize the tree as a leaf node

    While global stopping criteria are not met and the current tree can be further refined

    Do

    Pick the most promising leaf node to expand

    Draw sample of examples

    Use the trained network to label these examples

    Select a splitting test for the node

    For each possible outcome of the test make a new leaf node

    End

End

### 3.1 M-of-N Splitting tests

TREPAN uses the m-of-n test to partition the part of the instance space covered by a particular internal node. An m-of-n expression (a Boolean expression) is fulfilled when at least an integer threshold m of its n literals hold true. For example, consider four features a, b, c and d; the m-of-n test: 3-of-{a, b > 3.3, c, d} at a node signifies that if any of the 3 conditions of the given set of 4 are satisfied then an example will pass through that node. TREPAN employs a beam search method with beam width as a user defined parameter to find the best m-of-n test. Beam search is heuristic best-first each algorithm that evaluates that first n node (where n is a fixed value called the 'beam width') at each tree depth and picks the best out of them for the split. TREPAN uses both local and global stopping criteria. The growth of the tree stops when any of the following criteria are met: the size of the tree which is a user specific parameter or when all the training examples at node fall in the same class.

### 3.1 Single Test TREPAN and Disjunctive TREPAN

In addition to TREPAN algorithm, Craven has also developed two of its important variations. The single test TREPAN algorithm is similar to TREPAN in all respects except that as its name suggests it uses single feature tests at the internal nodes. Disjunctive TREPAN on the other hand, uses disjunctive "OR" tests at the internal nodes of the tree instead of the m-of-n tests. A more detailed explanation of the TREPAN algorithm can be found in Craven's dissertation [27].

Baesens et al [28] have applied TREPAN to credit risk evaluation and reported that it yields very good classification accuracy as compared to the logistic regression classifier and the popular C4.5 algorithm.

## 4. C4.5 ALGORITHM

The C4.5 algorithm [29] is one of the most widely used decision tree learning algorithms. It is an advanced and incremental software extension of the basic ID3 algorithm [30] designed to address the issues that were not dealt with by ID3. The C4.5 algorithm has its origins in Hunt's Concept Learning Systems (CLS) [31]. It is a non-incremental algorithm, which means that it derives its classes from an initial set of training instances. The classes derived from these instances are expected to work for all future test instances. The algorithm uses the greedy search approach to select the best attribute and never looks back to reconsider earlier choices. The C4.5 algorithm searches through the attributes of the training instances and finds the attribute that best separates the data. If this attribute perfectly classifies the training set then it stops else it recursively works on the remaining in the subsets (m = the remaining possible values of the attribute) to get their best attribute. Some attributes split the data more purely than others. Their values correspond more consistently with instances that have particular values of the target class. Therefore it can be said that they contain more information than the other attributes. But there should be a method that helps quantify this information and compares different attributes in the data which will enable us to decide which attribute should be placed at the highest node in the tree.

### 4.1 Information Gain, Entropy Measure and Gain Ratio

A fundamental part of any algorithm that constructs a decision tree from a dataset is the method in which it selects attributes at each node of the tree for splitting so that the depth of the tree is the minimum. ID3 uses the concept of Information Gain which is based on Information theory [32] to select the best attributes. Gain measures how well a given attribute separates training examples into its target classes. The one with the highest information is selected. Information gain calculates the reduction in entropy (or gain information) that would result from splitting the data into subsets based on an attribute.

The information gain of example set S on attribute A is defined as,

$$Gain(S, A) = Entropy(S) - \sum \frac{|S_v|}{|S|} Entropy(S_v) \qquad \text{Eq.1}$$

In the above equation, S is the number of instances and |Sv| is a subset of instances of S where A takes the value v. Entropy is a measure of the amount of information in an attribute. The higher the entropy, the more the information is required to completely describe the data. Hence, when building the decision tree, the idea is to decrease the entropy of the dataset until we reach a subset that is pure (a leaf), that has zero entropy and represents instances that all belong to one class. Entropy is given by,

$$Entropy(S) = \sum - p(I) \log_2 p(I) \qquad \text{Eq.2}$$

where p(I) is the proportion of S belonging to Class I.

Suppose we are constructing a decision tree with ID3 that will enable us to decide if the weather is favorable to play football. The input data to ID3 is shown in table 2 below adapted from Quinlan's C4.5.

Table 2. Play Tennis Examples Dataset

| Day | Outlook | Temperature | Humidity | Wind | Play Tennis |
|---|---|---|---|---|---|
| 1 | Sunny | Hot | High | Weak | No |
| 2 | Sunny | Hot | High | Strong | No |
| 3 | Overcast | Hot | High | Weak | Yes |
| 4 | Rain | Mild | High | Weak | Yes |
| 5 | Rain | Cool | Normal | Weak | Yes |
| 6 | Rain | Cool | Normal | Strong | No |
| 7 | Overcast | Cool | Normal | Strong | Yes |
| 8 | Sunny | Mild | High | Weak | No |
| 9 | Sunny | Cool | Normal | Weak | Yes |
| 10 | Rain | Mild | Normal | Weak | Yes |
| 11 | Sunny | Mild | Normal | Strong | Yes |
| 12 | Overcast | Mild | High | Strong | Yes |
| 13 | Overcast | Hot | Normal | Weak | Yes |
| 14 | Rain | Mild | High | Strong | No |

In this example,

$$Entropy(S) = -\left(\frac{9}{14}\right) \log_2 \left(\frac{9}{14}\right) - \left(\frac{5}{14}\right) \log_2 \left(\frac{5}{14}\right) = 0.9450 \qquad \text{Eq.3}$$

(Note: Number of instances where play tennis = yes is 9 and play tennis = No is 5)

The best attribute of the four is selected by calculating the Information Gain for each attribute as follows,

$$Gain(S, Outlook) = Entr.(S) - \frac{5}{14} Entr.(Sunny) - \frac{4}{14} Entr.(Overcast) - \frac{5}{14} Entr.(Rain) \qquad \text{Eq.4}$$

$$Gain(S, Outlook) = 0.9450 - 0.3364 - 0 - 0.3364 = 0.2670$$

Similarly, $Gain(S, Temp) = -0.42$ and $Gain(S, Wind) = 0.1515$

The attribute outlook has the highest gain and hence it is used as the decision attribute in the root node. The root node has three branches since the attribute outlook has three possible values, (Sunny, Overcast, and Rain). Only the remaining attributes are tested at the sunny branch node since outlook has already been used at the node. This process is recursively repeated until: all the training instances have been classified or every attribute has been utilized in the decision tree. The ID3 has a strong bias in favor of tests with many outcomes. Consider an employee database that consists of an employee identification number. Every attribute intended to be unique and partitioning any set of training cases on the values of this attribute will lead to a large number of subsets, each containing only one case. Hence the C4.5 algorithm incorporates use of a statistic called the "Gain Ratio" that compensates for the number of attributes by normalizing with information encoded in the split itself.

$$GainRatio = \frac{Gain(S, A)}{I(A)} \qquad \text{Eq.5}$$

In the above equation,

$$I(A) = \sum - p(I_A) \log_2 p(I_A) \qquad \text{Eq.6}$$

C4.5 has another advantage over ID3; it can deal with numeric attributes, missing values and noisy data.

## 5. EXPERIMENTATION AND RESULT ANALYSIS

We analyze three datasets with classes greater than two and we compare the results of Single-test TREPAN and C4.5 with that of X-TREPAN in terms of comprehensibility and classification accuracy. A generalized feed forward network was trained in order to investigate the ability of X-TREPAN in comprehending GFF networks. The traditional 'using-network' command was used to validate that X-TREPAN was producing correct outputs for the network. In all the experiments, we adopted the Single-test TREPAN as the best variant for comparison with the new model.

### 5.1 Body Fat

Body Fat is a regression problem in the simple machine learning dataset category. The instances are sought to predict body fat percentage based on body characteristics. A 14-4-1 MLP with hyperbolic tangent function was used to train the network for 1500 epochs giving an r (correlation co-efficient) value of 0.9882. Figure 6 shows the comparison of classification accuracy of body fat by the three models.

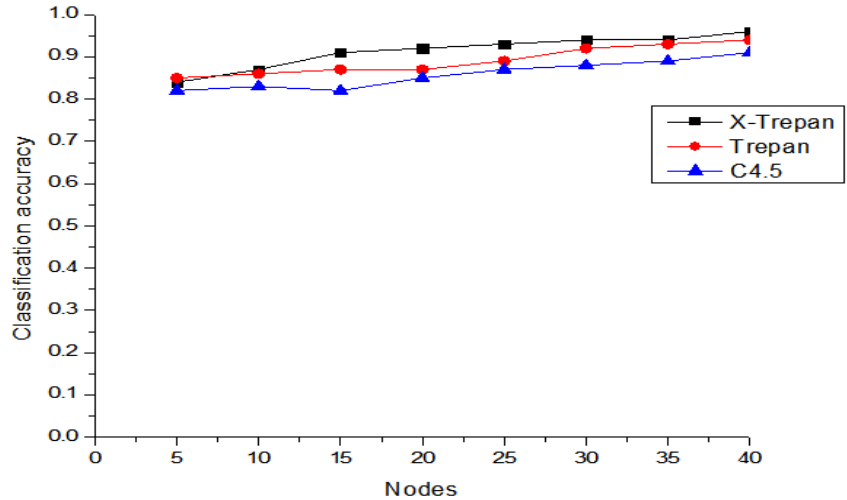

Figure 6. Comparison of classification accuracy of Body fat by the three algorithms

TREPAN achieves a classification accuracy of 94% and C4.5 produces a classification accuracy of 91% while X-TREPAN achieves a comparatively much higher accuracy of 96%. Additionally, both X-TREPAN and TREPAN generate similar trees in terms of size but accuracy and comprehensibility attained by X-TREPAN are comparatively higher.

The tables below show the confusion matrix of the classification accuracy achieved by TREPAN in comparison with X-TREPAN. While TREPAN produces a classification accuracy of 92.06% X-TREPAN produces a comparatively much higher accuracy of 96.83% as indicated in Table 3 below.

Table 3. Body Fat Confusion Matrix (X-TREPAN)

| **Actual/Predicted** | **Toned** | **Healthy** | **Flabby** | **Obese** |
|---|---|---|---|---|
| Toned | 13 | 0 | 0 | 0 |
| Healthy | 1 | 21 | 0 | 0 |
| Flabby | 0 | 0 | 9 | 1 |
| Obese | 0 | 0 | 0 | 18 |
| Classification Accuracy (%) | 92.86% | 100.00% | 100.00% | 94.74% |
| Total Accuracy (%) | | 96.83% | | |

Table 4. Body Fat Confusion Matrix (TREPAN)

| **Actual/Predicted** | **Toned** | **Healthy** | **Flabby** | **Obese** |
|---|---|---|---|---|
| Toned | 13 | 0 | 0 | 0 |
| Healthy | 1 | 20 | 0 | 0 |
| Flabby | 0 | 0 | 9 | 0 |
| Obese | 0 | 0 | 3 | 16 |
| Classification Accuracy (%) | 92.86% | 95.24% | 75.00% | 100.00% |
| Total Accuracy (%) | | 92.06% | | |

Additionally, both TREPAN and X-TREPAN generate identical trees in terms of size but accuracy attained by X-TREPAN is comparatively higher.

## 5.2 Outages

Outages constitute a database from the small dataset category. A12-3-1 MLP network with a hyperbolic tangent and bias axon transfer functions in the first and the second hidden layer respectively gave the best accuracy. The model was trained for 12000 epochs and achieved an r (correlation co-efficient) value of 0.985 (or an r2 of (0.985)2). Figure 7 shows the comparison of classification accuracy of outages by the three algorithms.

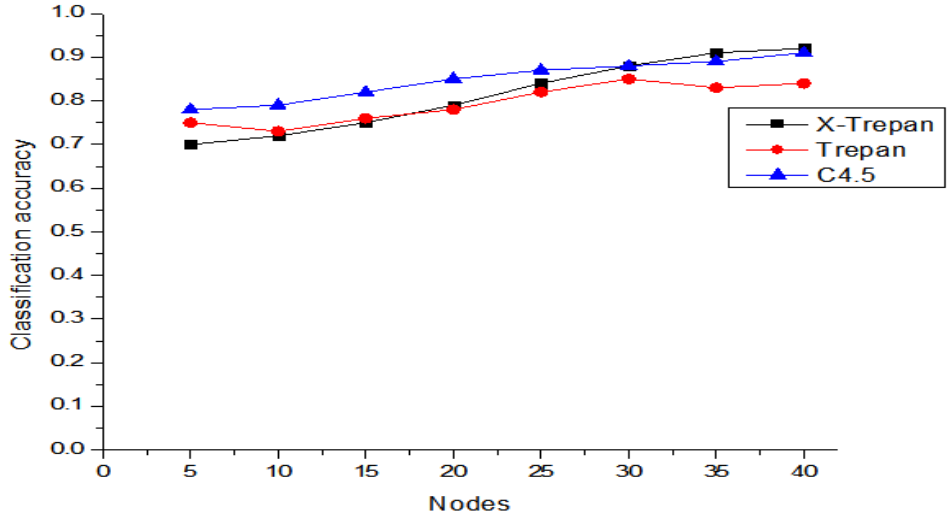

Figure 7. Comparison of classification accuracy of Outages by the three algorithms

In terms of classification accuracy, as can be seen in the figure above, TREPAN achieves 84%, C4.5 achieves 91% while X-TREPAN achieves 92%.

However, here TREPAN, C4.5 and X-TREPAN all generate very different trees in terms of size with C4.5 producing the largest and most complex decision tree while X-TREPAN produces the simplest and smallest decision tree with comparatively higher accuracy and comprehensibility.

The tables below show the confusion matrix of the classification accuracy achieved by both algorithms. X-TREPAN achieves 85% while TREPAN achieves 76%.

Table 5. Outages Confusion Matrix (X-TREPAN)

| Actual/Predicted | C11 | C12 | C13 | C14 | C15 |
|---|---|---|---|---|---|
| C11 | 3 | 0 | 0 | 0 | 0 |
| C12 | 4 | 48 | 5 | 0 | 0 |
| C13 | 0 | 1 | 8 | 0 | 0 |
| C14 | 0 | 0 | 1 | 5 | 0 |
| C15 | 0 | 0 | 0 | 0 | 0 |
| Classification Accuracy (%) | 42.86% | 97.96% | 57.14% | 100% | 0.00% |
| Total Accuracy (%) | | | 85.33% | | |

Table 6. Outages Confusion Matrix (TREPAN)

| Actual/Predicted | C11 | C12 | C13 | C14 | C15 |
|---|---|---|---|---|---|
| C11 | 2 | 5 | 0 | 0 | 0 |
| C12 | 3 | 43 | 3 | 0 | 0 |
| C13 | 0     | 6     | 7     | 1     | 0     |
| C14 | 0     | 0     | 0     | 5     | 0     |
| C15 | 0     | 0     | 0     | 0     | 0     |
| Classification Accuracy (%) | 40.00% | 79.63% | 70.00% | 83.33% | 0.00% |
| Total Accuracy (%) | | | 76.00% | | |

### 5.3 Admissions

A typical University admissions database model based on a 22-15-10-2 MLP network. Two hidden layers with the hyperbolic tangent transfer functions were used for modeling. The best model was obtained by the X-TREPAN with a minimum sample size of 1000, tree size of 50 and classification accuracy of 74%. Figure 8 gives the comparison of classification accuracy of Admissions by the three models.

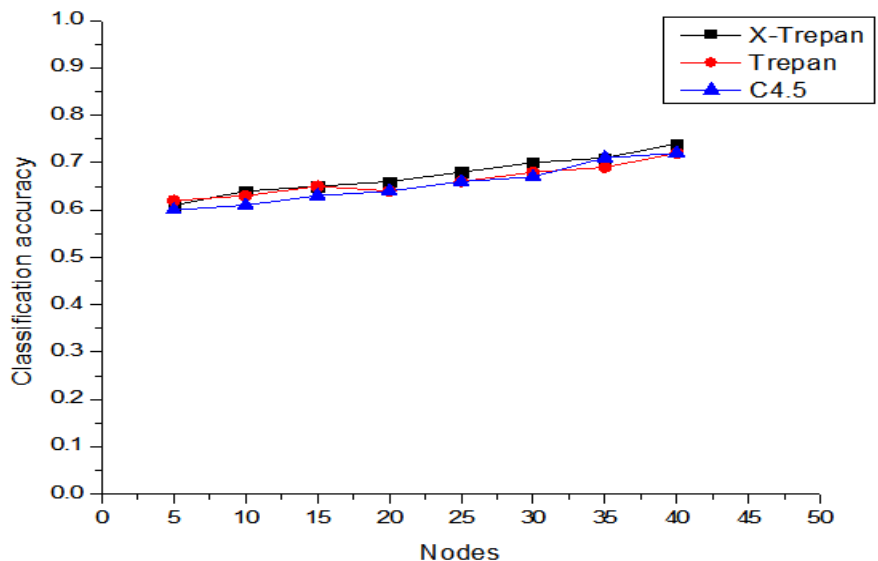

Figure 8. Comparison of classification accuracy of Admissions by the three algorithms

On the other hand, C4.5 achieved an accuracy of 71.97% (not rounded) almost equaling that of TREPAN of 72%, but produced a significantly large and complex decision tree.

In terms of Confusion Matrix, TREPAN achieved an accuracy of 71.6% very close to that of X-TREPAN. The confusion matrix is shown in the Tables below.

Table 7. Admissions Confusion Matrix (X-TREPAN)

| Actural/Predicted | Yes | No |
| --- | --- | --- |
| Yes | 401 | 279 |
| No | 168 | 754 |
| Classification Accuracy (%) | 70.47% | 72.99 |
| Total Accuracy (%) | 72.10% | |

Table 8. Admissions Confusion Matrix (TREPAN)

| Actural/Predicted | Yes | No |
| --- | --- | --- |
| Yes                        | 379    | 190    |
| No                         | 259    | 774    |
| Classification Accuracy (%) | 59.40% | 80.29  |
| Total Accuracy (%)         |     71.67%      ||

## 6. PERFORMANCE ASSESSMENT

### 6.1 Classification Accuracy

The classification accuracy or error rate is the percentage of correct predictions made by the model over a data set. It is assessed using the confusion matrix. A confusion matrix is a matrix plot of predicted versus actual classes with all of the correct classifications depicted along the diagonal of the matrix. It gives the number of correctly classified instances, incorrectly classified instances and overall classification accuracy.

The accuracy of the classifier is given by the formula,

$$Accuracy(\%) = \frac{(TP + TN)}{(TP + FN + FP + TN)} \times 100 \qquad \text{Eq.5}$$

Where true positive = (TP), true negative = (TN) false positive = (FP) and false negative = (FN). A false positive (FP) is when a negative instance incorrectly classified as a positive and false negative (FN) is when a positive instance is incorrectly classified as a negative. A true positive (TP) is when an instance is correctly classified as positive and true negative (TN) is when an instance is correctly classified as negative and so on.

A confusion matrix is a primary tool in visualizing the performance of a classifier. However it does not take into account the fact that some misclassifications are worse than others. To overcome this problem we use a measure called the Kappa Statistic which considers the fact that correct values in a confusion matrix are due to chance agreement.

The Kappa statistic is defined as,

$$\hat{k} = \frac{P((A) - P(E))}{1 - P(E)} \qquad \text{Eq.6}$$

In this equation, P(A) is the proportion of times the model values were equal to the actual value and, P(E) is the expected proportion by Chance.

For perfect agreement, Kappa = 1. For example: a Kappa statistic of 0.84 would imply that the classification process was avoiding 84% of the errors that a completely random classification would generate.

### 6.2 Comprehensibility

The comprehensibility of the tree structure decreases with the increase in the size and complexity. The principle of Occam's Razors says "when you have two competing theories which make exactly the same projections, the one that is simpler is the better" [33]. Therefore, among the three algorithms, X-TREPAN is better as it produces smaller and simpler trees as against Single-test TREPAN and C4.5 in most scenarios.

# 7. CONCLUSION

The TREPAN algorithm code was modified (X-TREPAN) to be able to work with multi-class regression type problems. Various experiments were run to investigate its compatibility with generalized feed forward networks. The weights and network file were restructured to present GFF networks in a format recognized by X-TREPAN. Neural Network models were trained on each dataset varying parameters like network architecture and transfer functions. The weights and biases obtained from the trained models of the three datasets were fed to X-TREPAN for decision tree learning from neural networks. For performance assessment, classification accuracy of Single-test TREPAN, C4.5 and X-TREPAN were compared. In the scenarios discussed in the paper, X-TREPAN model significantly outperformed the Single-test TREPAN and C4.5 algorithms in terms of classification accuracy as well as size, complexity and comprehensibility of decision trees. To validate the results, we use classification accuracy not as the only measure of performance, but also the kappa statistics. The kappa statistical values further validate the conclusions that X-TREPAN is a better one in terms of decision tree induction.